\documentclass[a4paper]{article}

\usepackage[english]{babel}
\usepackage[utf8]{inputenc}
\usepackage{amsmath}
\usepackage{graphicx}
\usepackage[colorinlistoftodos]{todonotes}

\usepackage{authblk}
\usepackage{url}
          
\usepackage{pgfplots, pgfplotstable}
\pgfplotsset{
    single xbar legend/.style={
        legend image code/.code={\draw[##1,/tikz/.cd,bar width=6pt,bar shift=0pt,xbar] plot coordinates {(0.8em,0pt)};},
    }
}

\title{Unraveling reported dreams with text analytics}
\author[1,2]{Iris Hendrickx} 
 \author[1]{Louis Onrust}
 \author[1]{Florian Kunneman}
\author[1]{Ali Hürriyetoğlu}
\author[1,2]{Antal van den Bosch}
  \author[2]{Wessel Stoop}

\affil[1]{Centre for Language Studies,
Radboud University Nijmegen, The Netherlands
}
\affil[2]{Centre for Language and Speech Technology, Radboud University Nijmegen, The Netherlands}

\date{}

\begin{document}
\maketitle

\begin{abstract}
We investigate what distinguishes reported dreams from other personal narratives.
The {\it continuity hypothesis}, stemming from psychological dream analysis work, states that most dreams refer to a person's daily life and personal concerns, similar to other personal narratives such as diary entries. Differences between the two texts may reveal the linguistic markers of dream text, which could be the basis for new dream analysis work and for the automatic detection of dream descriptions.
We used three text analytics methods: text classification, topic modeling, and text coherence analysis, and applied these methods to a balanced set of texts representing dreams, diary entries, and other personal stories.
We observed that dream texts could be distinguished from other personal narratives nearly perfectly, mostly based on the presence of uncertainty markers and descriptions of scenes. Important markers for non-dream narratives are specific time expressions and conversational expressions. Dream texts also exhibit a lower discourse coherence than other personal narratives.
\end{abstract}

\section{Introduction}




The analysis of dreams has a long history. One of the earliest recorded dream analyses was written on clay tablets in Mesopotamia, 5000 years ago \cite{Black+1992}. This ancient epic tale of Gilgamesh includes several dream descriptions and interpretations. In ancient Greek and Egyptian times, dreams were seen as messages from the gods. Despite the various research fields that study the meaning and purpose of dreams, such as psychiatry, psychology, neuroscience, and religious studies, a comprehensive explanation of the purpose of dreams is still lacking.

Psychologists and social scientists have studied dream content with quantitative methods for decades, working with the hypothesis that dreams reveal psychological information about the dreamer. One currently dominant theory in this area is the {\it continuity hypothesis}, which assumes that the content of dreams reflects a person's daily life and personal concerns \cite{domhoff1996finding}. Previous studies on dream descriptions, i.e.\ reported dreams written afterwards by the dreamer, have shown that around 75--80\% of dream content relates to everyday settings, characters, and activities. The remaining dreams are related to uncommon or even bizarre topics. Some of the latter are shared by numerous people, such as dreaming about flying, teeth falling out, or being naked in public \cite{dh2008db}. 
 
 
Dream descriptions are written reports of the memories of an experienced dream. Even though much progress is made in brain studies, it is not possible yet to decode the dream content from a dreaming person's brain activity. The only possible way to gather dream contents is to study the reported recollection produced when the experiencer was awake \cite{domhoff1996finding}. For this reason, we study written reports of remembered dreams. As a textual genre, this type of written report bears similarities to other written recollections of personal experiences, both in cognitive and sensory qualities \cite{Kahan2011}. In this study we aim to investigate what linguistic features are specific to dream reports, contrary to reports on personal stories that actually happened, using tools for automatic text analysis. 
Computational approaches to automatically analyze the content of dream reports from a linguistic perspective are rare. In this work we want to pave the road for further detailed and knowledge-directed research by presenting a first account of a computational text analysis of dream reports.

We performed three different types of automatic text analysis to investigate what typical characteristics we can discover in dream reports. We hope that our automatic linguistic approach can demonstrate to dream analysis experts how well-studied techniques from the field of computational linguistics can be applied to offer insights into linguistic patterns hidden in large dream collections.

The largest available digitally curated collection of dream reports is the DreamBank \cite{dh2008db} which contains over 22 thousand dream reports gathered in the last decades in various scientific  studies. We use DreamBank as the base for our study; we also collected a contrasting data sample of true personal stories (from Reddit and Prosebox) to perform our experiments. 

We apply the following three methods: {\it automatic text classification}\/ to investigate what features are actually salient for predicting whether a written text is a dream report or not, {\it topic modeling}\/ to discover the common themes in the dream collection, and {\it text coherence analysis}\/ to measure whether there is a difference in coherence between dreams and personal stories.
Each of these methods offers a different perspective on dream data. As we will argue, they do lead to overlapping findings that we discuss in the last section. 

This paper is structured as follows. We first discuss related work in automatic textual analysis of dream reports and previous work on comparisons between dreams and stories in Section \ref{sec-rel-work}. We present the data sets used in the experiments in detail in Section \ref{sec-data}. Next we present the three different studies we have done in Section \ref{sec-methods}, and we summarize and discuss our findings in Section \ref{sec-end}.

\section{Related work}\label{sec-rel-work}

Automatic textual analysis of dream reports is a relatively unexplored field. Semi-automatic experiments have been performed by Bulkeley and Domhoff \cite{bulkeley2009seeking}, who developed a systematic category list of word strings that can be used for automated queries and word-frequency counts. The categories in which the words are organized relate to the content of dreams, and are used to count mentions of emotions, characters, perception, movement, cognition, and culture. In a more recent follow-up study \cite{Bulkeley2014} this category list is updated and evaluated on four data sets present in the DreamBank corpus. The study shows that one can use this type of word analysis to detect the general topics in dream content in the same way and with a similar accuracy as the more time-consuming manual analysis. Furthermore, Bulkeley offers evidence that based on an individual dream collection, it is possible to make accurate estimations about a person's life, his concerns, activities, and interests, thereby confirming the continuity hypothesis.

Some work exists on automatic text classification with machine learning methods, where the task is to assign emotion labels to dreams. In \cite{Ravazi2014} (follow up work on \cite{Matwin2010}) the authors aim to label dreams on a four-point negative/positive sentiment scale. The authors represent dreams as word vectors and include dynamic features to represent sentiment changes in the dream story. They run ten-fold cross validation experiments on a sample of 477 manually labeled dream reports and achieve up to 64\% accuracy, close to the average human agreement of 69\%. 
%

In \cite{frantova2009automatic} a more refined type of sentiment analysis is explored; they predict the fuzzy assignment of five emotion categories to dream descriptions, based on semi-automatically compiled emotion word dictionaries. Their method is evaluated against a sample from the DreamBank that is manually labeled with the emotion annotations from the Hall/Van de Castle encoding system \cite{hall1966content}. The difficulty of using these DreamBank annotations is that this labeling has been done at the document level, which is also the level at which \cite{frantova2009automatic} evaluated their approach, even though the annotations refer to specific phrases in the dream. The direct link between the linguistic description and label is missing. 

In previous non-computational studies dreams have been compared to stories from a narrative perspective. To what extent dreams can be considered a story or narrative is highly dependent on the exact definition of a story or narrative, as argued by Kilroe \cite{Kilroe2000}. Simply put, a narrative or story is the report of a sequence of events that takes place in a certain setting and involves one or more characters. The causality of the events and the way the events in a pattern, is called a plot. Not every story has to have a plot \cite{Forster1956}.
A more specific definition is given by Montagero \cite{montangero2012dreams} 
who specifies that the characters in a narrative need to have intentional states and that a narrative must introduce an unexpected event. He argues that dreams indeed can be classified as narratives under this definition.

Drawing on the argumentation of this previous work we posit that dreams are personal narratives, which narrows down our research question to: what makes dream reports different from other personal narratives such as true stories?

\section{Data}\label{sec-data}
As we are interested in automatically investigating textual properties, and studying what characteristics are typical for dream reports, we compare dream reports to other texts and narratives. We use dream reports from the DreamBank, and we place them in contrast with data representing personal narratives that actually happened, taken from the internet sources Reddit and Prosebox. In this section we introduce the three sources, and describe their properties.

\subsection{DreamBank}\label{sec-DreamBank}

We use the dream reports from collections as gathered in the DreamBank, a project to combine the results of several scientific studies and resources over the years in one online search interface \cite{dh2008db}. These collections of dream series vary greatly in type, size, and intended purpose. Some series consist of a longitudinal collection of dream descriptions of a single person, such as the collection ``Dorothea: 53 years of dreams'' consisting of around 900 dreams. Other series represent a specific group of dreamers such as adult male and female blind dreamers \cite{blinddreamers1999}. Some collections are in English, gathered in Australia, Canada or the US. One of the collections, collected in Switzerland, contains dream reports in German. The DreamBank is an ongoing project and collections are added regularly. We use a snapshot from the DreamBank retrieved in April 2015 containing 22 thousand dreams divided in 67 different collections. 

For our experiments we performed the following selection steps on the DreamBank data, where we limited ourselves to English written collections. Since some of the DreamBank collections overlap, we removed the duplicates from our sample. We also removed a part of the description in the collection ``College women from the late 1940'' that contained answers to specific questions, and we only kept the dream description itself. We applied an automatic language identification step \cite{lui2012langid}\footnote{We used langid.py version 1.1.5 (github hash: e801bf8, accessible at \url{http://git.io/vcc2Z}).} that filtered out a handful of other dreams (for example dream \#0694 of the Barb Sanders collection \cite{domhoff2006barb} where she dreams about a Spanish conversation). Next, the data was tokenized automatically,\footnote{We used twokenize.py, which is part of twitter\_nlp (github hash 27c8190, accessible at \url{http://git.io/vccyu}).} leading to a sample of 21,598 dream descriptions containing a total of 4.3 million word tokens. Dream descriptions contain an average of 56 words, with a population standard deviation of 38.5. 

We noted that some collections in the DreamBank are much larger than others, and that dream descriptions of certain persons (e.g.\ Barb Sanders) are relatively prominent in the DreamBank content. We decided to create a sample of the DreamBank where we limit the amount of dream reports per individual dreamer to a random selection of at most one hundred dreams. This produced a sample of 6,998 dream descriptions, comprising 1.3 million tokens in total with an average of 65 words and a standard deviation of 43.7 per dream description, very similar to the larger sample. We used both the large and the small sample in our experiments. We show an example of a tokenized dream description with 97 tokens:

\begin{quotation}
\noindent I was chosen to be interviewed by S , the college president , but it's unclear if my papers were approved in time , so I clutch my briefcase with my acceptance letter in it and try to find the building . \\
A woman student and Ellie help guide me to the building . \\
I find a sign saying `` 504 , '' the room . \\
I rush to the room , hoping , feeling late and uncertain . \\
I am there in the nick of time . \\
I am calm and handle it well . \\
\end{quotation}

\subsection{Personal Stories}

To discover what the typical linguistic attributes of dream reports are, we need a comparable set of contrasting reports that is as similar to dream reports as possible, both in structure and in content. Comparing dream reports to a collection of news paper articles or personal letters will lead to obvious findings such as: dreams do not report on political debates and the weather forecast, and will not end in `yours sincerely'. This is not the type of differences that we are interested in. We therefore aimed to find a collection of personally written recollections of true daily life events. Recall that dreams are known to reflect daily life events and activities for at least 75--80\% of the cases.

Comparable collections of personal stories recollecting true events, not just fantasies or fiction, are difficult to find when looking for existing curated corpus collections. For this reason we resorted to collections of web texts to build our own corpus.

The first part of the contrasting data consists of personal stories. The stories are crawled from Prosebox,\footnote{\url{https://www.prosebox.net/}} an online community to share journals and personal stories. Just prior to this research, OpenDiary, a community where users could post diary entries, was taken down. Many of these users moved to Prosebox, and especially older posts are mostly diary entries or journals.

We collected all public posts that were available at the end of March 2015. As a result, we crawled 130 thousand posts with over 67 million tokens. We applied the same filtering pipeline to the Prosebox posts as was applied to the dreams; that is, we applied a language filter where we only kept the posts which were identified as English; second, we tokenized the posts. Since the number of tokens is much larger, we downsampled the corpus into a similar number of tokens as the DreamBank samples, i.e.\ the large sample and the smaller limited sample, containing 4.3 million words and 1.3 million words respectively, with an average of 64 and 63 words and standard deviations 78.8 and 94.3, respectively. In other words, we kept the average document size virtually equal to that of the DreamBank samples; the Prosebox data does exhibit a larger variance in size.
We show an excerpt of a Prosebox text here:
\begin{quotation}
\noindent Just sitting \\
Life is good here . \\
I had a good day of just staying home yesterday . \\
I went grocery shopping this morning and Cap is at his auction . \\
He has called me a couple of times and he is having a great time . \\
He loves seeing his friends that he sits with . \\
Tomorrow should be another day of staying home . \\
Yay . \\
If I lived by myself I wouldn't go anywhere . \\
I love staying home . \\
I bought groceries today . \\
I bought strawberries and whipping cream for strawberry shortcakes . \\
\end{quotation}

The second part of the contrast data consists of Reddit posts. Reddit\footnote{\url{https://www.reddit.com/}} is a website where users can submit content of almost every kind. The site uses a community system, where each community is called a subreddit. We collected posts from a number of subreddits where the posts are texts about daily and personal experiences such as communities named {\it offmychest}, {\it relationships}, {\it depression}, {\it lifeinapost}, and {\it self}. Prior to this research, the complete Reddit  corpus was not available.\footnote{See \url{https://redd.it/3bxlg7} for a dataset with all 1.7 billion publically available Reddit posts.} In total we crawled 122 thousand posts with 54 million words with an average post length of 71 words for the 1.3 million words sample, and 72 words for the 4.3 million words sample, respectively with standard deviations of 61.7 and 67.8.  We show an example Reddit story here:

\begin{quotation}
\noindent Down in my hole . \\
I am `` down in my hole '' my PTSD/ depression us usually `` under control '' but the past few days it feels like I am in molasses , all I want to do is sleep . \\
I want to cry , I want to run head long into a wall . \\
I will not hurt my self no matter how much I want to . \\
I just don't know what to do . \\
\end{quotation}

\section{Experiments}\label{sec-methods}

We applied topic modeling, text classification, and coherence tests to the aforementioned data sets in order to compare them.

\subsection{Text Classification Experiments}

As a first analysis of the text collection, we set out to train machine-learning classifiers to distinguish dream reports from personal stories. 
In text classification, a machine learning classifier is fed with labeled documents from which it learns to model the characteristics of the given labels. Its labeling performance is tested by applying the classifier on a held out set of documents. For this experiment, we used the sets of 4.3 million words for both the dream data and contrasting data.\footnote{We ran the same experiments on the small sets too and found virtually the same results.}

We tokenized all documents with the Stanford Tokenizer.\footnote{\url{http://nlp.stanford.edu/software/tokenizer.shtml}} The word tokens were standardized to lowercase. We extracted word unigrams, bigrams, and trigrams as features. To avoid bias from explicit markers of dream reports, we removed any features that contained one of the following words: dream, dreamer, dreamt, dreamed, dreams, awake, awaken, woke.

We compared the performance of three different classification algorithms: Support Vector Machines (SVM), Naive Bayes, and Balanced Winnow. We used the libsvm \cite{Hsu+03} implementation of SVM, with linear kernel and setting the C parameter to $1.0$. We applied Naive Bayes by using the Multinomial Naive Bayes implementation in Scikit-learn \cite{scikit-learn}.\footnote{\url{http://scikit-learn.org/stable/modules/naive_bayes.html}} For Balanced Winnow, we made use of the Linguistic Classification System \cite{koster2003multi}. 
The $\alpha$ and $\beta$ parameters were set to $1.05$ and $0.95$ respectively. The major threshold ($\theta+$) and the minor threshold ($\theta-$) were set to $2.5$ and $0.5$. The number of iterations was bound to one. 

We evaluated the performance of the three approaches by means of ten-fold cross-validation. To avoid author bias, the reports by the same author were kept together in either the test set or the train set during each fold. During each training phase, the 7,500 most frequent features were selected and presented as binary values. 

The classification results, micro-averaged over examples, are given in Table \ref{tab:performance}. All three approaches yield a precision and recall of $0.97$, which indicates that the dream and non-dream reports can easily be distinguished with a small remaining margin of error. Table \ref{tab:performance} also displays the exact number of documents that were correctly classified. The Balanced Winnow classifier has a slightly higher number of correct classifications than Naive Bayes and SVM. 

\begin{table}
\begin{tabular}{l|lllllll}
Approach & Prec & Recall & F1 & TPR & FPR & AUC & \# correct\\  
& & & & & & & \\ \hline
SVM & 0.97 & 0.97 & 0.97 & 0.97 & 0.03 & 0.97 & 38,225 \\ 
Balanced Winnow & 0.97 & 0.97 & 0.97 & 0.97 & 0.03 & 0.97 & 38,334 \\ 
Naive Bayes & 0.97 & 0.97 & 0.97 & 0.97 & 0.03 & 0.97 & 38,229 \\
\end{tabular}
\caption{Micro-averaged performance of three classifiers on distinguishing dream reports from reports of real-world events on the 4.3M words corpus (39,480 documents in total). TPR = True Positive Rate. FPR = False Positive Rate. AUC = Area Under the Curve, \# correct = number of correctly labeled documents.}
\label{tab:performance}
\end{table}


The Balanced Winnow classifier returns an interpretable model of the features that the classifier used internally to make its predictions. Upon analysis of the 30 most indicative features of the dream and non-dream classes, we obtained the following insights about the two types of texts:

\begin{itemize}
	\item Dream reports are characterized by words that convey uncertainty: somebody (rank 3), remember (rank 5), somewhere (rank 12) and recall (rank 17);
    \item Another category of features that have a high rank in dream reports are references to a space or situation: setting (rank 1), riding (rank 8), building (rank 16), swimming (rank 23), table (rank 25) and room (rank 30);
    \item In contrast to dream reports, personal stories contain indications of specific points in time: 2014 (rank 4), today (rank 9), tonight (rank 19), yesterday (rank 21), day (rank 23) and months (rank 28);
    \item Personal stories are also distinguished by conversational utterances, such as `:)' (rank 2), please (rank 17), `?' (rank 20) and thanks (rank 27).  
\end{itemize}


\subsection{Topic modeling}

To discover what type of topics are typical for dream reports, we applied an unsupervised method that is currently popular for discovering latent themes or topics in large document collections. Latent Dirichlet Allocation (LDA) \cite{blei2012} is a probabilistic generative algorithm that aims to give a broad overview of the topics that occur in a collection of documents. Topics are defined as a distribution over a fixed vocabulary (in practice, a topic consists of a set of related words). Topic modeling is an unsupervised process solely based on word occurrences in documents. LDA assumes that documents are created based on an underlying topic distribution and each document is generated from a mixture of these topics that each have a different proportion in the document. LDA uses an iterative process to estimate this underlying distribution based on the observed words in the text. 



We ran experiments with LDA on the full DreamBank sample of 22,046 dreams. We filtered the dream texts to exclude all function words and punctuation marks and only kept those content words that were automatically part-of-speech tagged by the Stanford parser as nouns, verbs and adjectives. All words have been converted to lower case. Such explicit filtering step ensures that the generated LDA topics contain only content words.

For these experiments we use the LDA implementation provided in the Mallet toolkit \cite{mccallum2002mallet}. We ran LDA with 2,000 iterations with Gibbs sampling and 50 topics. The produced LDA model was used to annotate each document with its most relevant topics; namely those topics that cover at least 10 percent of the document. Documents have three such topics on average.

Setting the number of topics parameter is a rather arbitrary choice. We ran experiments with 100, 200 and 400 topics as well and studied the output. When raising the value of this parameter, more fine-grained topic descriptions are produced. These detailed topics are still understandable and coherent topics, but, as can be expected, they tend to have a lower coverage in the document. 
As we aim to look at significant differences between topic distributions in different sample sets and to compute g-tests (log-likelihood tests) \cite{rayson2000comparing}, we choose to keep the number of topics fixed at 50.

\subsubsection{LDA on the DreamBank}

LDA can give surprising insights in the data. We applied LDA to the full DreamBank set of dreams and we present a random representative sample of these topics in Figure \ref{tab-lda-dream}. The number in each row denotes the topic number and does not express a ranking or weight.  Certain topics express a specific script or frame; in the first three topics in Figure \ref{tab-lda-dream} we see `purchasing', `using the bathroom' and `school life'.
Other topics express a setting such as `inside the house' in topic 42, and the 'outdoor' setting in topic 5. It is also remarkable to see how narrative verbs are clustered together in present tense in topic 35 and in past tense in topic 48. These verbs are commonly used in action and event descriptions (do, say, see).
These automatically generated topics are a clear support for the {\it continuity hypothesis} \cite{domhoff1996finding} as they reflect daily life events, characters and settings.

\begin{table}
\begin{tabular}{ll}

44 & money pay get give buy bank bill machine change\\
37 & bathroom water toilet shower use clean bath floor sink\\
25 & class school teacher students high test room classroom college\\
42 & room door house see window open apartment go living\\
5 & road hill tree see walking snow mountain trees people\\
28 & love feel kiss make happy want man other hug\\
35 & say says do see go man woman comes get\\
48 & said did went came got told started saw looked asked\\

 \end{tabular}  
 \caption{\label{tab-lda-dream} Examples from the topics generated on the DreamBank sample.}
 \end{table}
 
In a next step we zoom in on two comparable dream sets of men and women to study the differences in topics between these groups. We use the normative male and female dream sample present in the DreamBank (abbreviated to  {\it Hall/VdC Norms}) based on the older work of Hall and Van de Castle \cite{hall1966content}. 

Topics were generated based on the full sample. For each topic we compute whether the topic occurs significantly more or less in {\it Hall/VdC Norms} male dreams than female dreams. We computed a g-test\footnote{We used the g-test implementation written by Pete Hurd, available at \url{http://www.psych.ualberta.ca/~phurd/cruft/g.test.r}} on the frequency topic counts with $p<0.05$. In Figure~\ref{tab-lda-mf} we show the topics that occurred significantly more in either the male or the female dream sub-sample. Remarkable stereotypical differences can be found in the topics; men tend to dream more about shooting, driving, sex, and games, while women dream more about weddings, fashion, and family.

The interpretation of the topics is not always straightforward as the machine views documents does not view the text in the way humans do, and sometimes the resulting topics can be harder to interpret \cite{chang2009reading}, such as topics 11 or 39, which both seem to grasp a less clear topic related to dreaming in general.

These preliminary results are in line with recent research on differences between male and female in \cite{domhoff2015correcting}, and other work such as \cite{schredl2005gender} and \cite{dh2008db}, which states ``[that] there are more appearances of tools and cars in men's dreams, more appearances of clothing and household items in women's dreams''. The main difference is that the results presented here were obtained unsupervised, and support the current manually found results reported in other papers.


\begin{table}
\begin{tabular}{ll}
0  & gun fire men shot shoot man police shooting war deer\\
5  & road hill tree see walking snow mountain trees people side\\
11 & dream remember seemed do time being other same feeling something\\
15 & car driving drive road street get truck going front side\\
17 & bed room sleep sex sleeping lying bedroom floor naked lay\\
29 & game playing ball play team basketball football field baseball cards\\
33 & plane fly flying sky air see airplane land people ground\\
34 & building floor stairs get go elevator people steps see walking\\
\hline \\
12 & wedding married john wife getting ring husband george bonita ceremony\\
14 & things room put stuff box small old boxes take find\\
21 & wearing white dress black blue dressed clothes red shirt shoes\\
23 & house mother father home brother family old sister parents children\\
26 & get go do going trying take want find time know\\
39 & girl friend dream girls friends remember went dreamed did school\\
40 & man woman men young other women boy old older small\\
 \end{tabular}  
 \caption{\label{tab-lda-mf} Male (top) and female (bottom) topics in the DreamBank. \label{tab-topicmf} }
 \end{table}

 \subsubsection{LDA on dreams and stories}

In the next step we combined the dream sample with the Reddit and Prosebox samples into one large collection on which we ran the LDA topic modeling using the same setting of 50 topics.

To investigate what topics occur significantly more with dreams than with personal stories, we took a random sample of 2,000 dreams and 2,000 stories and computed a g-test to check whether there were significant differences in the topic distributions. In this sample of 4,000 documents we found that 42 of the 50 topics occur significantly more or significantly less with either dreams or personal stories. This shows that there are clear differences between the two sets; more so than between the male and female dreams. In Table \ref{tab-lda-dn} we show the top five most significantly different topics for dreams and stories.
 
Topic 28 is typical for what we expect to be a dream description, mentioning words such as `dream', `remember' and `seemed'. We observe an `inside the house' setting description (topic 23) and an aquatic setting description (topic 1). The other two topics express related narrative description verbs in present and paste tense.

For the personal stories we observe two topics that are directly linked to the Reddit categories that were included in the sample, namely `anxiety' and `relationships'. Topic 45 expresses conversational internet language including profanities and abbreviated verb forms. Topic 24 consists mostly of time expressions.

There is some overlap in the most important words in the topic word lists. The term `get' occurs in the top words lists of four of the five most significant personal stories topics.  The terms `see' and `saw' occur in respectively three and two of the significant dream topics.

\begin{table}
\begin{tabular}{ll}

23 & room house door see go floor open stairs window apartment\\
40 & saw came said went looked got ran walked horse did\\
1  & water see boat pool river swimming lake beach go people\\
28 & dream remember seemed girl man boy came saw dreamed being\\
13 & see go says say man woman get look comes walk\\
\hline \\
26 & do time get things think going know something make other\\
45 & fucking shit fuck do get im know day got dont\\
30 & feel do life help depression anxiety get know feeling want\\
22 & relationship do want feel other months boyfriend tl girlfriend friends\\
24 & day last today work get going week night got time\\
 \end{tabular}  
 \caption{Five most significant topics for dreams (top) and non-dreams (bottom). \label{tab-lda-dn} }
 \end{table}

 \subsection{Bizarreness as dream characteristic}
 
When people are asked what is typical about dreams, they will often mention weirdness as a typical property of dreams. In dreams strange events or weird things seem to occur. This might be due to the fact that most dreams are forgotten the next morning and only weird or impressive dreams stick to people's memory \cite{bulkeley2011big}. This recollection could be attributed to the bizarreness effect, the inclination to remember bizarre things better than ordinary things \cite{mcdaniel1995bizarreness}. 
As Domhoff shows in \cite{domhoff2007realistic} bizarreness does occur in dreams, but it is not as manifest as people tend to believe. 
 
Interestingly, bizarre thoughts do not only occur in our sleep but can also occur when awake. It has been shown that people in a relaxed undisturbed awake condition produce dream-like reports when asked to recall that most recent thoughts in the same way subjects are asked after being awakened \cite{reinsel1992waking}. These awake fantasies are very similar to dream reports including bizarreness. In the study of Reinsel bizarreness was measured by counting three different type of occurrences: discontinuous events, improbable combinations, and improbable entities. Discontinuous events were found to be the most contributing factor in bizarreness (around 60\% of the counts) while improbable entities were much less present (only around 8\%).
This is in line with previous studies on large volumes of dream reports that show that the amount of bizarreness attributed to strange entities is relatively low; most characters in dreams are known persons \cite{dh2008db}. 
 
 
One specific type of bizarreness is metamorphosis.
Domhoff \cite{domhoff2003scientific} (p.132) investigated metamorphoses via a search in DreamBank.net and found only 50 mentions of metamorphoses in the whole DreamBank. Transformations turn out to be highly infrequent in dreams.
\cite{Schweickert+2010} study metamorphosis as a typical dream phenomenon and focus on the relation between change in form to change in inner states. No evidence was found that form change is connected to a change in mental state. This was a small-scale study on a set of 65 dreams from 21 persons. 

In our study we aimed at using a quantitative approximation of bizarreness and applying this metric to the DreamBank texts. We focus on the discontinuity of events in dreams and try to quantify this by looking at textual coherence in the dream reports. We hypothesize that dreams are less coherent in their discourse structure than personal stories. We measure two different aspects of discourse structure, namely discourse marker frequency and entity-based text coherence, using the smaller balanced sample sets of 1.3 million words.

Discourse analysis is a broad and multi-disciplinary field that studies language in use beyond the sentence level \cite{trappes2004}.
Automatic discourse parsing is still in its early development phase as was illustrated by in this year's CoNLL shared task \cite{xue-EtAl:2015:CoNLL-ST} on shallow discourse parsing where the best system achieved a overall F-score of 24\%. 

In this initial experiment we only focus on discourse marker occurrences and measure whether there is a difference between discourse marker frequency in the dream data and the personal stories. Discourse markers are words or phrases that explicitly signal discourse structure and describe how two sentences or phrases are related to each other. For example, {\it for example}\/ indicates that the current sentence exemplifies something that was mentioned in the previous sentence. Typical discourse markers are {\it but}, {\it since}, {\it while}, {\it even though}, and {\it because}.

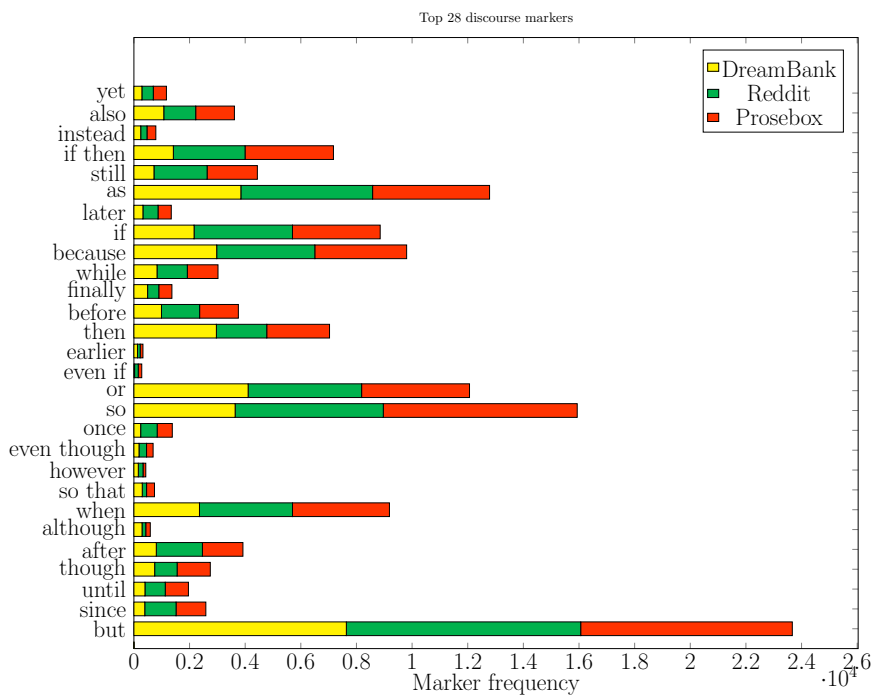
\begin{figure}[htb]
\resizebox{0.95\textwidth}{!}{%
\begin{tikzpicture}
\begin{axis}[
    xbar stacked,   
    xmin=0,         
    ymin=0,
    xlabel = Marker frequency,
    ytick=data,     
    bar width=10,
    title= Top 28 discourse markers,
    yticklabel style = {font=\LARGE,xshift=-0.5ex},
    xticklabel style = {font=\LARGE},
    legend style = {font=\LARGE},
    xlabel style = {font=\LARGE},
    yticklabels={,but,since,until,though,after,although,when,so that,however,even though,once,so,or,even if,earlier,then,before,finally,while,because,if,later,as,still,if then,instead,also,yet},
    width=20cm
]
\addplot [fill=yellow] table [x=First, y expr=\coordindex] {data1.csv};    
\addplot [fill=green!70!blue]table [x=Second, y expr=\coordindex] {data1.csv};
\addplot [fill=red!80!yellow] table [x=Third, y expr=\coordindex] {data1.csv};
\legend{\strut DreamBank,Reddit,Prosebox}
\end{axis}
\end{tikzpicture}
}%

\caption{Frequency of discourse markers per dataset and their total number. Discourse markers used have a frequency of more than 250 in the Penn Discourse Treebank. The `then' marker, a typical discourse marker used in sequential narration, occurs substantially more often in dreams than in Prosebox or Reddit posts.}
\label{fig-markersav}
\end{figure}

We used a list of 60 markers\footnote{We excluded `and' as discourse marker in this experiment due to its ambiguity as conjunction marker and high frequency.} based on annotations from the Penn Discourse Treebank \cite{prasad2008penn} that was used in the CoNLL shared task. In total, about 40 thousand occurrences of these markers were counted in the DreamBank data, and about 50 thousand in both Reddit and Prosebox, which means that there are about 20\% fewer discourse markers counted in the DreamBank data than in the contrasting data. In Figure \ref{fig-markersav} we show the distribution of 28 of the 60 markers in the balanced DreamBank, Reddit and Prosebox data sets, having a frequency of occurrence over 250.
For most markers we can observe similar distributions, or slightly lower count for the dreams. One marker however occurs substantially more often with the dreams than with Prosebox or Reddit: `then'. This is a typical discourse marker that is used in sequential narration.

In a second experiment we study entity-based coherence. Mentioned entities and chains of referring expressions in a text are core indicators of text coherence. We assume that discontinuity in dreams is expressed in sudden shifts in scenes and events, and we expect that these are linked to shifts in mentioned entities. On the basis of this assumption, we tried to measure discourse coherence by applying an existing automatic model to detect entity-based coherence.

We used the Brown Coherence Toolkit v1.0  \cite{elsner2008coreference}.
The authors of this toolkit present an extension of the entity-grid coherence model proposed by Barzilay and Lapata \cite{barzilay2008modeling}. An entity-grid represents the entity mentions in a document in a textual matrix where each row represents an entity and the column represent the syntactic roles of the entities (subject, object, other). This matrix is used to predict which role each entity will have in the next sentence.

To detect the entities in the text we used the extended entity grid  based on the Wall Street Journal corpus that was automatically pre-processed with OpenNLP software,  available in the Brown Coherence Toolkit.

We applied the model to each of the balanced dream and personal stories data sets and measure its performance with a binary discrimination test as was previously done in the work of Elsner and Charniak \cite{elsner2008coreference}. The binary discrimination test tests the model's ability to distinguish between a human-authored document in its original order, and a random permutation of that document. The test reads any number of documents and performs the test on each one, using 20 random permutations.


The results of this test are shown in Table \ref{tab-brown}. All achieved results are substantially lower than the scores reported by Elsner and Charniak who report scores of 86\% F-score when training on Wall Street Journal (WSJ) newspaper text and testing on a held-out set from the same corpus. As WSJ consist of financial news paper articles, we can expect a drop in performance when switching to a completely different textual genre of dreams and personal stories. Nevertheless, the scores in Table~\ref{tab-brown} indeed suggest that the dream text is less coherent than more formal edited text in terms of coherence relations, but also as compared to Prosebox and Reddit, which are remarkably similar. 

 \begin{table}
 \begin{center}
\begin{tabular}{lll}
       Dataset & Accuracy & F-score \\
       \hline
        DreamBank & 0.23 & 0.32  \\
        Prosebox & 0.37 & 0.42  \\
        Reddit & 0.37 & 0.43  \\
\end{tabular}
\caption{\label{tab-brown} Results from the entity-based coherence model as evaluated by a binary discrimination test.  }
\end{center}
\end{table}

\section{Discussion}\label{sec-end}

We presented three automatic text analysis studies of dream reports. First we performed a supervised text classification experiment to see how easy or hard it is to distinguish dream reports from texts that are closely related in both content and structure, namely true personal stories. We applied three different text classification algorithms to the same task; they all succeeded in labeling the documents with a near-perfect precision. Differentiating between dreams and personal stories turned out an easy task. The analysis of the features used by the Balanced Winnow classifier show that expressions of uncertainty and setting descriptions and narrative verbs are typical for dreams, while time expressions and conversational expressions are typical for the personal stories.

In the second study we aimed to explore the general topics that are present in the full DreamBank. We applied LDA topic modeling to the DreamBank to study the main themes in dreams. The results mostly showed topics describing everyday activities, settings, and characters.
This unsupervised method signaled the same differences as the text classification experiments between dreams and stories: setting descriptions and uncertainty expressions are typical for dreams while the use of time expressions and conversational expressions occur more often in stories.
In our exploratory study on discontinuity in dreams, we saw that dream reports indeed use less discourse markers and have a lower entity-based textual coherence. With these experiments we are only just scratching the surface of doing automatic discourse analysis but we feel that these preliminary experiments are a starting point for further analyses in this direction.


Even though our experiments have shown some interesting and consistent findings about the typical differences between dreams and stories, we need to be careful with our conclusions. The fact that the text classifiers obtained such high scores and the topics were significantly different distributed over the samples, can be an indication that the contrasting data sample was not as representative as we had hoped for.
The emerging topic about `anxiety' for example shows that this subreddit had a substantial influence on the results. 
We suspect that a more careful selection over a much larger set of personal stories, and perhaps an additional check to filter out characteristic internet language is needed to create automatic models that focus on the more subtle differences between reported dreams and personal experiences from real life.

As a next step, we plan experiments on another sample of personal stories and dreams to investigate the effect of the sample representativeness. 
We also aim to collaborate with dream analysis experts to work towards better interpretations of the results that we found and to explore further research questions in the area of dream analysis.

Furthermore, we are interested in the question whether humans can just as easily distinguish between dream descriptions and true stories as the text classifier could. We are currently working on building an online human judgment task to investigate this question.



\section*{Acknowledgments}
 We would like to thank Kelly Bulkeley and G.\ William Domhoff for their valuable feedback and suggestions.
We also thank G.\ William Domhoff and Adam Scheider for creating the Dreambank that was the underpinning for this study.

\bibliographystyle{alpha}
\bibliography{dreams}


\end{document}